\documentclass[letterpaper, 10 pt, journal, twoside]{IEEEtran}

\usepackage[table,xcdraw]{xcolor}

\usepackage{kotex}
\usepackage{graphicx}
\usepackage{tabularx}
\usepackage{amsmath}
\usepackage{mathtools}
\usepackage{graphicx}
\usepackage{mathtools}
\usepackage{multirow}
\usepackage{hhline}
\usepackage{booktabs}
\usepackage[]{algorithm2e}
\usepackage{mathabx}
\usepackage{eucal}
\usepackage{txfonts}
\usepackage{color,soul}
\usepackage{array}
\usepackage{tikz}
\usepackage{comment}
\usepackage{amssymb} 
\usepackage{color}
\usepackage{cite}
\usepackage{arydshln}

\newcommand{\Tref}[1]{Table~\ref{#1}}
\newcommand{\Eref}[1]{Eq.~(\ref{#1})}
\newcommand{\Fref}[1]{Fig.~\ref{#1}}
\newcommand{\Sref}[1]{Section~\ref{#1}}

\title{
Learning Off-Road Terrain Traversability with Self-Supervisions Only
}

\author{Junwon Seo, Sungdae Sim, and Inwook Shim
\thanks{Manuscript received: February, 7, 2023; Revised May, 4, 2023; Accepted May, 29, 2023. This paper was recommended for publication by Editor V. Markus upon evaluation of the Associate Editor and Reviewers' comments. This work was supported by the Agency For Defense Development Grant funded by the Korean Government in 2023. {(Corresponding author: Inwook Shim.)}} 
\thanks{Junwon Seo and Sungdae Sim are with the Agency for Defense Development, Daejeon 34186, Republic of Korea
        {\tt\footnotesize \{junwon.vision, sungdae.sim\}@gmail.com}}%
\thanks{Inwook Shim was with the Agency for Defense Development, Daejeon 34186, Republic of Korea. He is now with the Department of Smart Mobility Engineering, Inha University, Republic of Korea
        {\tt\footnotesize iwshim@inha.ac.kr}}%
\thanks{The multimedia material is available at \texttt{\small{https://bit.ly/3YdKanw}}.}%
\thanks{Digital Object Identifier (DOI): see top of this page.}
}
\begin{document}

\markboth{IEEE Robotics and Automation Letters. Preprint Version. ACCEPTED MAY, 2023}
{Seo \MakeLowercase{\textit{et al.}}: Learning Off-Road Terrain Traversability with Self-Supervisions Only} 
\maketitle

\begin{abstract}
Estimating the traversability of terrain should be reliable and accurate in diverse conditions for autonomous driving in off-road environments. However, learning-based approaches often yield unreliable results when confronted with unfamiliar contexts, and it is challenging to obtain manual annotations frequently for new circumstances. In this paper, we introduce a method for learning traversability from images that utilizes only self-supervision and no manual labels, enabling it to easily learn traversability in new circumstances. To this end, we first generate self-supervised traversability labels from past driving trajectories by labeling regions traversed by the vehicle as highly traversable. Using the self-supervised labels, we then train a neural network that identifies terrains that are safe to traverse from an image using a one-class classification algorithm. Additionally, we supplement the limitations of self-supervised labels by incorporating methods of self-supervised learning of visual representations. To conduct a comprehensive evaluation, we collect data in a variety of driving environments and perceptual conditions and show that our method produces reliable estimations in various environments. In addition, the experimental results validate that our method outperforms other self-supervised traversability estimation methods and achieves comparable performances with supervised learning methods trained on manually labeled data.
\end{abstract}


\section{INTRODUCTION}
\IEEEPARstart{R}{ecent} advancements in visual perception enabled the success of fast-moving autonomous off-road vehicles. Estimating the traversability of the terrain with visual sensors is a crucial component of off-road driving. Numerous studies have made significant improvements in traversability estimation using large-scale datasets with human annotations and RGB images that provide semantically rich information about complex environments~\cite{geiger2013vision, Jiang_RELLIS3D}.  However, the datasets only contain observations for a specific and limited context, resulting in unreliable estimates for unobserved conditions. 

To successfully adapt to new circumstances, frequent manual annotation is required, which is not only unsustainable but also erroneous. Due to the high cost of the data-labeling procedure, obtaining sufficient labeled data regarding the various environments would be challenging. The labels produced by human experts often provide inadequate information for learning traversability in complex environments, since the ground truth regarding traversable regions can not be clearly defined in off-road environments. In addition, the domain-specific annotations would lose their relevance in unfamiliar environments. For instance, various conditions, including places, seasons, weather, lighting, and camera settings, can significantly affect the visual appearance of an outdoor environment and the performance of estimations. Consequently, the vehicle cannot accurately predict traversability from images in a variety of situations if only static and constrained datasets are utilized.

While it is impractical to manually annotate images of every single environment, labels on traversable regions can be automatically generated by exploiting the vehicle trajectories in a self-supervised fashion~\cite{wellhausen_2019should, Z_acoustic, Wellhausen_2020, schmid_2022self, hirose2018gonet, seo_2022scate}. Various works present the self-supervised approaches to learning traversability, which leverage self-supervised traversability labels instead of human-provided annotations ~\cite{wellhausen_2019should, Dahlkamp-Self, Z_acoustic, Wellhausen_2020, schmid_2022self, hirose2018gonet, seo_2022scate}. However, they focus mostly on traversal cost analysis or terrain categorization in confined contexts, rather than identifying traversable regions reliably in diverse environments. In order for a vehicle to operate successfully and sustainably in a range of environments, it would be desirable that traversability is learned solely by utilizing the self-supervised traversability data.

\begin{figure}[t]
\centering
\includegraphics[width=0.9\linewidth]{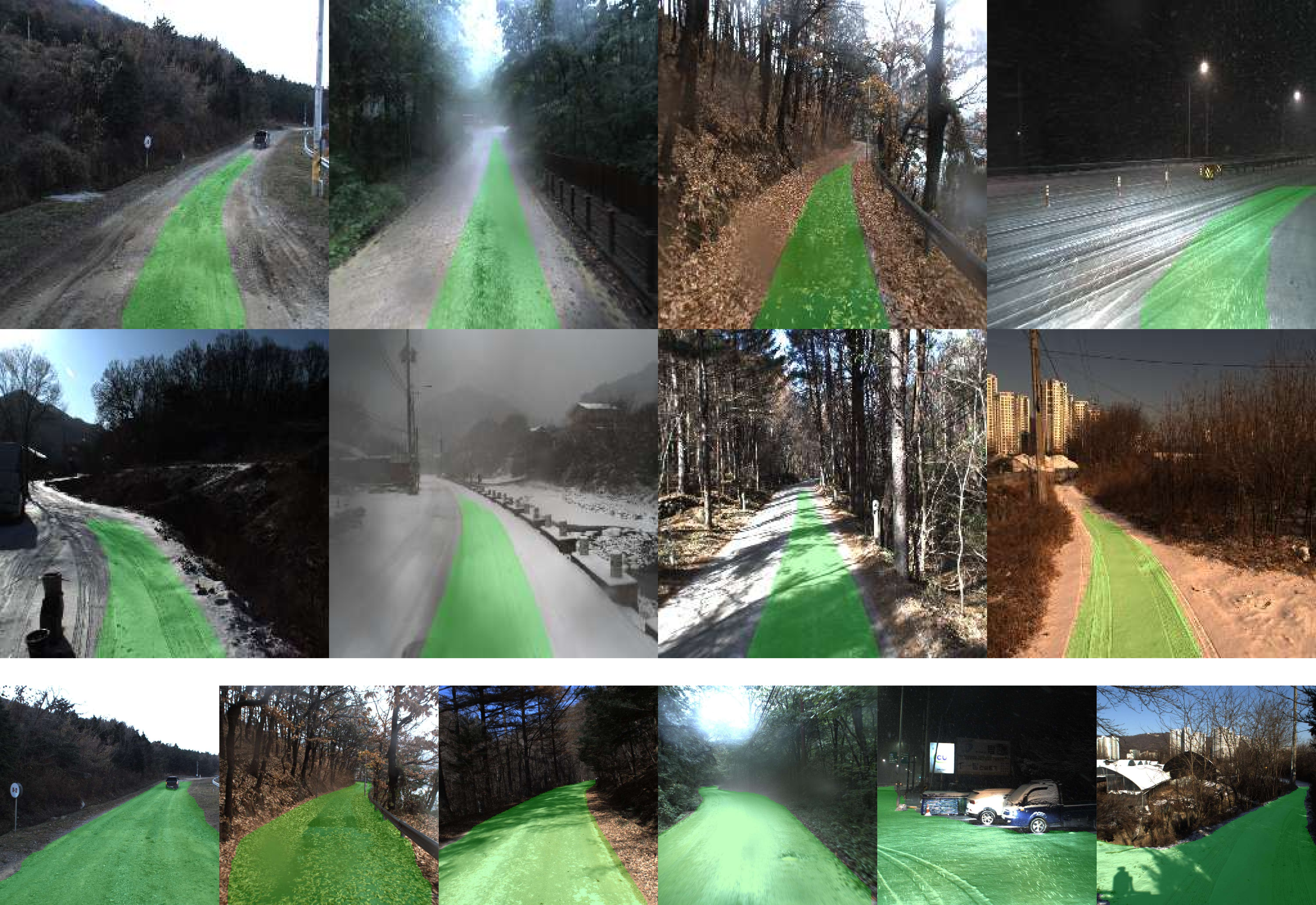}
\caption{We present a traversability estimation method that can be trained without human annotation in various environments. \textit{Top}: Self-supervised traversability data gathered on diverse environments. There exists a large variance in visual appearances. \textit{Bottom}: Traversability estimation results of the model learned solely with the self-supervised traversability labels.}
\label{fig:MAIN}
\end{figure}

Nonetheless, learning traversability from self-supervised traversability data is challenging due to the following reasons. First, since the vehicle has no experience traversing non-traversable regions, no labels for non-traversable regions can be obtained. While supervised learning-based methods\cite{zhao_2017PSPNET} learn to differentiate between regions with distinct labels, the self-supervised traversability data only contains labels for a single class. It lacks supervision for discriminating between traversable and non-traversable regions, leading to overconfident predictions~\cite{wellhausen_2019should}. Second, the self-supervised traversability label is incomplete. Only a small portion of the traversable regions are labeled by trajectories, leaving the remainder unlabeled, and some of the labels may be inaccurate due to occlusions of trajectories~\cite{seo_2022scate}. Lastly, while some methods leverage one-class classification methods for learning with the self-supervised label, they not only fail to produce a dependable prediction for off-road images but also do not conduct experiments in various environments.

In this work, we propose a self-supervised traversability estimation method that learns traversability only from self-supervisions without explicit labels. We present an automated labeling process that can produce reliable self-supervised traversability labels on images by utilizing past vehicle trajectories. With the self-supervised labels, our algorithm learns traversability in off-road environments with complex distributions by leveraging Positive-Unlabeled(PU)~\cite{bekker_2020PUlearning} learning method and the $2D$ normalizing flow~\cite{yu_2021fastflow}. Moreover, to complement for insufficient supervision of the self-supervised labels, we employ approaches for self-supervised learning of visual representations to obtain discriminative representations from images~\cite{chen_2020simclr}.

To demonstrate the efficacy of our method, we collect large-scale driving data under a variety of conditions, including terrain types, places, weather, seasons, and lighting conditions. We conduct extensive experiments using our dataset along with the public dataset, RELLIS-3D~\cite{Jiang_RELLIS3D}. Our comprehensive quantitative and qualitative evaluations demonstrate that our method can effectively learn traversability in a wide range of unstructured and unknown environments.

\section{RELATED WORKS}
\subsection{Traversability Estimation}
With developments in learning traversability, autonomous driving has made significant progress in urban and off-road environments~\cite{geiger2012_kitti}. Most early works on estimating traversability concentrate on analyzing simple geometric and visual features~\cite{Wermelinger_Navigation, sock_traversability}. With the development of deep neural networks, semantic segmentation is widely utilized to classify terrains into predefined terrain classes leveraging large datasets~\cite{borges_2022survey}. In addition, numerous approaches have been developed to identify traversable regions in unstructured environments~\cite{borges_2022survey, Shaban_TerrainClassification}. Fully convolutional networks for image segmentation~\cite{zhao_2017PSPNET} have significantly improved the off-road traversability estimation performance since images contain semantically rich and dense information about the off-road environments~\cite{guan2022ga}. However, such methods heavily rely on training data, which leads to incorrect estimations when confronted with data from distributions not included in the training data~\cite{barnes_2017findYourOwn}. The supervised learning methods may not generalize well to changing and unknown environmental circumstances~\cite{Z_acoustic}. For the widespread deployment of autonomous vehicles off-road, where the likelihood of encountering an unfamiliar context is considerable, the model should be capable of working reliably in various environments.

\subsection{Self-Supervised Learning of Traversability}
For reliable traversability estimation in a wide range of environments, self-supervised approaches are proposed, which exploit a vehicle’s driving experience to learn the traversability of a terrain~\cite{Dahlkamp-Self, kim_2006traversability, wellhausen_2019should, Z_acoustic, castro_2022howdoesitfeel}. These methods enable automated procedures to self-label visual data for learning traversability. For example, measurements from proprioceptive sensors are used to assess the traversal cost of terrain or to classify terrains. However, they either rely on manual labels or are oblivious to the fact that estimations in unseen environments can be unreliable. As labels on non-traversable regions cannot be acquired via self-labeling, the estimations are prone to over-confident predictions, which might lead to navigational failure~\cite{Wellhausen_2020}. Consequently, identifying traversable regions with reliability is a crucial problem. While our previous work~\cite{seo_2022scate} has shown that traversability can be learned using point clouds, the method for identifying the traversable region using images has not been exhaustively examined in a variety of environments.

One-class classification algorithms can be employed to distinguish traversable and non-traversable regions~\cite{Wellhausen_2020, pmlr-v155-ji21a, schmid_2022self, hirose2018gonet}. For example, normalizing flow~\cite{dinh_2017nvp} shows a great performance for traversability classification on multi-modal images~\cite{Wellhausen_2020}. However, it freezes the feature encoder after pretraining and encodes local patches, which do not incorporate global scene information. The features would simply capture low-level meanings, such as the color and texture of terrains, without their semantic information required to discriminate between traversable and non-traversable terrain~\cite{gudovskiy_2022cflow}. In addition, an autoencoder is used to identify high-risk terrains based on the reconstruction error~\cite{schmid_2022self}. The simple autoencoder-based reconstruction focuses on low-frequency details, resulting in the estimation that high-frequency details are simply classified as non-traversable. The autoencoder is also known to generalize well to unseen data, resulting in large false-positive predictions in which well-reconstructed non-traversable regions are assigned a low anomaly score~\cite{gong_2019memorizing}. The one-class classification algorithm should be capable of extracting more discriminative features from images with self-supervised labels in order to reliably identify traversable regions in diverse off-road contexts.

\subsection{Self-Supervised Learning of Visual Representations}
Instead of predicting human-annotated labels, approaches on self-supervised visual pre-training learn without labels by solving pretext tasks. The pretext tasks include the reconstruction of inputs, instance discriminations, and clustering with pseudo-labels~\cite{caron_2020unsupervised, asano_2020self}. Most state-of-the-art methods are contrastive learning, in which the network is trained to attract positive sample pairs and repel negative sample pairs~\cite{chen_2020simclr}. Due to the incomplete labeling of the self-supervised traversability data, the acquisition of highly discriminative features from the data is challenging. By complementing the short supervision of self-supervised traversability data with self-supervised learning of visual representations, the visual representation for learning traversability could be more discriminative.

\section{METHODS}
Our goal is to train a network that can successfully embed the complex data distribution of environments, allowing for precise and reliable estimation in a wide range of unstructured contexts. Given image $\mathbf{x}$, we generate a self-supervised label $\hat{\mathbf{y}}$ that does not require manual annotation. Then, we learn a model that estimates a pixel-wise traversability $\mathbf{y}_i = f(\mathbf{x}_i)$, where $\mathbf{x}_i$ is an image pixel and $\mathbf{y}_i$ is terrain traversability representing whether or not the terrain is traversable.

\subsection{Self-Supervised Traversability Label}\label{sec:label}
\begin{figure}[t]
\centering
\includegraphics[trim={0 1cm 0 0},clip,width=0.99\linewidth]{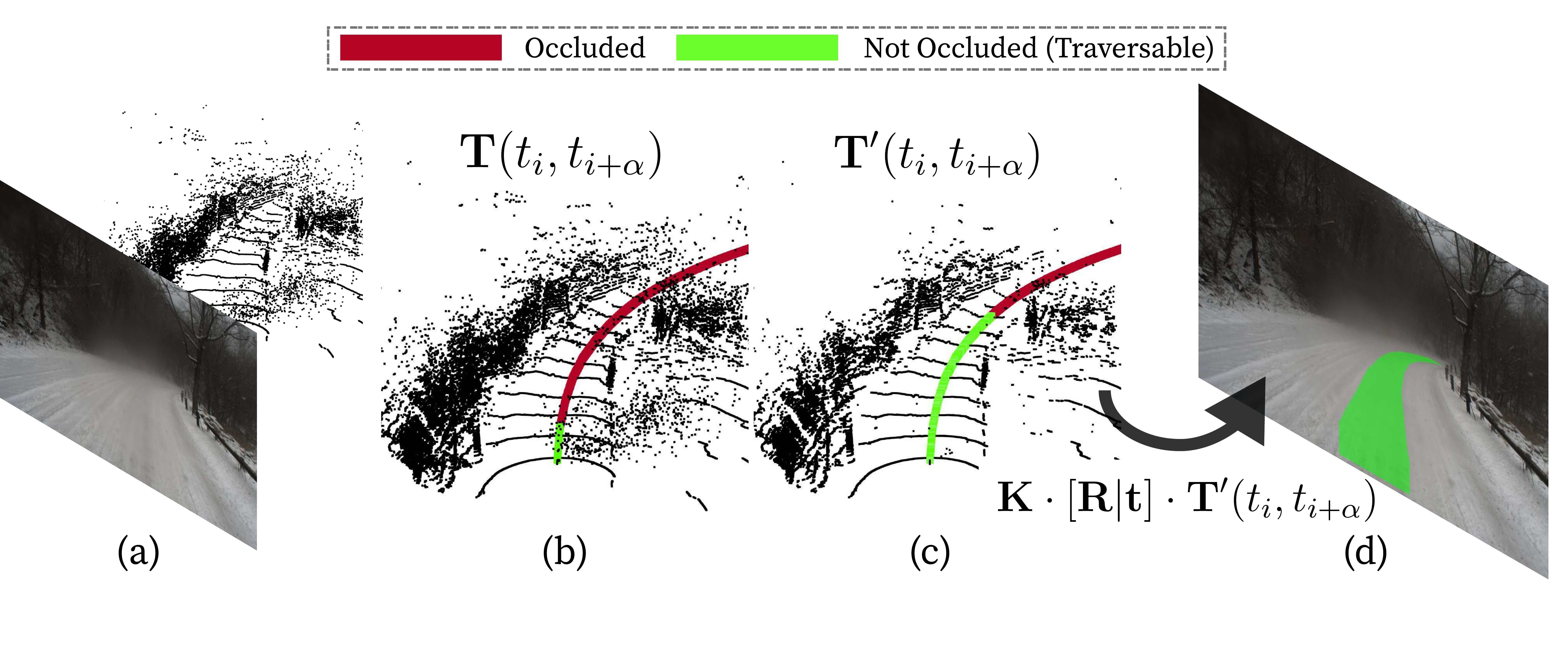}
\caption{Overview of our proposed automated procedure for self-supervised traversability label generation. (a) On captured data, the wheel-contact points are considered traversable. The occluded points are filtered before projection to image coordinates to eliminate false positive labels. (b) Noises in LiDAR points are regarded as obstacles, leading to erroneous filtering. (c) Unsupervised LiDAR denoising is performed to improve the efficacy of filtering in off-road conditions. (d) Trajectories are then projected to image coordinates and labeled as traversable.
}
\label{fig:Data_Generation}
\end{figure}

From images gathered while driving, the self-supervised label $\hat{\mathbf{y}}$ is generated by an automated procedure, as illustrated in \Fref{fig:Data_Generation}. Since the regions traversed by a vehicle during data collection can be considered safe to traverse, we can designate such regions as traversable. The wheel-terrain contact points are calculated using the trajectories recovered by the SLAM~\cite{slam}. The trajectories of horizon $\alpha$ from time $t_i$ are converted to contact points, denoted as $\mathbf{T}(t_{i}, t_{i+\alpha})$. 

Prior to the labeling, the contact points are filtered to eliminate false-positive labels. Since the past trajectory is projected onto the $2D$ images, parts of the contact points can be occluded due to obstacles or rotations of the vehicle. Without filtering, the obstacles would be labeled as traversable, leading to a large number of false positives in estimations. Although these false-positive labels can be avoided by shortening the horizons, this shortens supervision for learning. The contact points are therefore filtered using LiDAR points captured simultaneously with the images. Similar to an occlusion filtering algorithm~\cite{schmid_2022self}, a contact point is filtered as occluded if it has a longer radial distance than the nearest LiDAR points in spherical coordinates.


However, numerous undesirable noises exist in LiDAR point clouds acquired under a variety of off-road conditions~(e.g. dust, rain, snow). The noises can be regarded as an obstacle during filtering, which may hinder the effectiveness of filtering. For robust labeling in a variety of unstructured environments, unsupervised LiDAR denoising~\cite{bae_2022slide} is performed prior to occlusion filtering. With the denoised point cloud, the contact points are filtered, denoted as $\mathbf{T}'(t_{i}, t_{i+\alpha})$. 

Finally, the contact points are projected into the camera coordinates to generate the self-supervised label of the PU type by the following equations:
\begin{equation}\label{data_generation}
    \hat{\mathbf{y}} = \mathbf{K} \cdot [\mathbf{R}|\mathbf{t}] \cdot \mathbf{T}'(t_{i}, t_{i+\alpha})
\end{equation}
where $\mathbf{K}$ and $[\mathbf{R}|\mathbf{t}]$ represent the intrinsic camera calibration matrix and the world-to-camera transformation matrix respectively. On the image coordinates, pixels between the left and right wheel-terrain contact points are labeled as \textit{positive}, while all other pixels are left \textit{unlabeled}. Note that the label consists of a relatively small number of positive pixels and unlabeled pixels are a combination of traversable and non-traversable regions.

\subsection{Learning Traversability}
\begin{figure*}[t]
\centering
\includegraphics[width=0.9\linewidth]{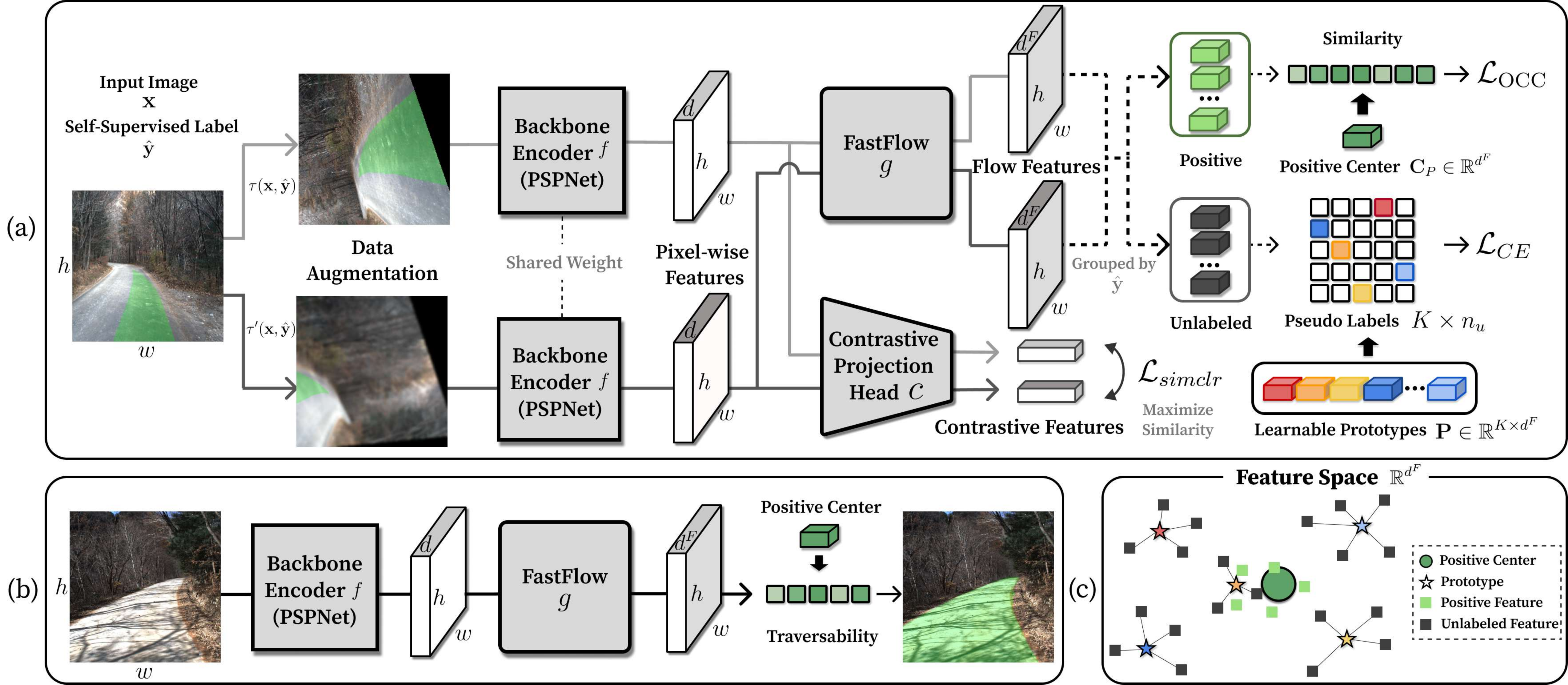}
\caption{High-level structure of the proposed method. (a) Training. (b) Inference. (c) Illustration of the embedding space of flow features at training. The feature space is learned with the clustering pretext task, enhancing the discriminative power of features for the one-class classification.
}
\label{fig:ARCHITECTURE}
\end{figure*}

In this section, we propose a method for learning traversability with self-supervised labels only. The overall architecture of our learning method is illustrated in \Fref{fig:ARCHITECTURE}. 

\subsubsection{One-Class Classification}
The image is forwarded to a feature encoder for image segmentation, denoted as $f$, that maps images $\mathbf{x} \in \mathbb{R}^{h \times w \times 3}$ into pixel-wise features $\mathbf{z} \in \mathbb{R}^{h \times w \times d}$. Our backbone encoder is PSPNet~\cite{zhao_2017PSPNET} which can capture global context through the pyramid pooling module. 

The feature embedding space can be trained to minimize the volume of a positive-data-enclosing hypersphere. Then, the similarity metric between a feature and the center of the hypersphere, $p(\mathbf{z}_i) \in [0, 1]$, can be used to determine the traversability of the $i^{th}$ pixel. The simple one-class classification loss can be used for positive pixels:
\begin{equation}\label{eq:svdd}
    \mathcal{L}_{\text{OCC}} = 1 - p(\mathbf{z}_i).
\end{equation}

However, the representations of positive pixels have high intra-class variation as there exist various representations of traversable regions in off-road environments. The loss function~\Eref{eq:svdd} assumes a single center and pushes dissimilar features towards a single center, thereby diminishing the discriminative power of the representations. Not only is the solution susceptible to a hypersphere-collapse solution, in which the majority of data can be trivially mapped to the hypersphere center, but it is also incapable of effectively capturing multimodal distributions.

Normalizing flow~\cite{dinh_2017nvp} can be used to project complex distributions of features to a simple distribution while avoiding a trivial solution. The flow model, denoted as $g$, transforms a pixel-wise feature $\mathbf{z}_i \in \mathbb{R}^{d}$ into a flow feature $\mathbf{z}^F_i \in \mathbb{R}^{d^F}$ with a tractable distribution using a bijective invertible mapping. We adopt the $2D$ normalizing flow model with affine coupling layers, Fastflow~\cite{yu_2021fastflow}, which produces more accurate features for segmentation. The likelihood of a flow feature can be simply defined as $p(\mathbf{z_i}^F) = \mathbf{z_i}^F \cdot \mathbf{C}_p$, which represents cosine similarity with the hypersphere center of positive data, $\mathbf{C}_p \in \mathbb{R}^{d^F}$. Then, the traversability of an image pixel can be easily calculated using the change of the variable formula:
\begin{equation}\label{pos_score}
\begin{aligned} 
\log {p}(\mathbf{z}_i)  &= \log p(\mathbf{z}^F_i) + \log \left\vert \mathbf{det}(\frac{\partial \mathbf{z}^F_i}{\partial \mathbf{z}_i}) \right\vert,
\end{aligned}
\end{equation}where the determinant of the Jacobian $\frac{\partial \mathbf{z}^F_i}{\partial \mathbf{z}_i}$ can be calculated with affine coupling layers~\cite{dinh_2017nvp}. Intuitively, the Jacobian penalizes the trivial solutions that have constant mappings. 

However, the normalizing flow model cannot be end-to-end trained with a feature encoder because it would generate a trivial backbone encoder while preventing the trivial flow model. In addition, the network does not utilize unlabeled data. The network would be overfitted to the distributions of traversable regions, limiting its ability for discriminating between traversable and non-traversable regions.

\subsubsection{Self-Supervised Clustering with Unlabeled Data}
We use unlabeled data in a self-supervised manner to train the network end-to-end so that the network can learn better embeddings while avoiding trivial solutions. Motivated by the clustering-based self-supervised learning of visual representations~\cite{caron_2020unsupervised, asano_2020self}, our methodology solves the clustering pretext task. The flow features of unlabeled pixels are jointly self-labeled and clustered with a set of $K$ learnable prototypes, $\mathbf{P} \in \mathbb{R}^{K \times d^F}$, which functions as cluster centers. 

By taking the softmax of the similarity between prototypes and unlabeled features, the posterior distribution of unlabeled pixels to prototypes, $\mathbf{Q} \in \mathbb{R}^{K \times n_u}$, is computed, where $n_u$ is the number of unlabeled pixels within a batch. Then, soft cluster assignment $\mathbf{A} \in \mathbb{R}^{K \times n_u}$ from features to prototypes is computed by optimizing the following equation with an equipartition constraint:
\begin{equation}\label{assignment}
    \max_{\mathbf{A}} \operatorname{Tr}(\mathbf{A}^\intercal \mathbf{Q}) \quad s.t. \quad \mathbf{A}\cdot\mathbf{1}^{n_u} = \frac{n_u}{K}\cdot\mathbf{1}^{K}
\end{equation}

The constraints ensure that the prototypes equally partition the assignments, thereby preventing trivial solutions in which features are collapsed into equal representations~\cite{caron_2020unsupervised}. This optimization problem can be efficiently solved with a few iterations of the \textit{Sinkhorn-Knopp} algorithm~\cite{sinkhorn}. By minimizing the cross-entropy loss between the posterior distribution and the optimized cluster assignment, the features and prototypes are simultaneously updated:
\begin{equation}\label{CE}
    \mathcal{L}_{\text{CE}} = - \frac{1}{n_u} \sum_k^{K}\sum_j^{n_u} \mathbf{A}_{kj} \log (\mathbf{Q}_{kj}).
\end{equation} 

However, the learned representations are still insufficient because the supervision of the self-supervised labels is restricted to a small portion of the entire traversable regions. In contrast to supervised learning methods, which are trained to explicitly distinguish traversable and non-traversable regions with full labels, the unsupervised clustering objective may attempt to learn simplistic features.

\subsubsection{Self-Supervised Contrastive Learning}
To supplement the representation power of backbone features, the encoder $f$ is simultaneously learned with the contrastive pretext task of self-supervised visual representation learning~\cite{chen_2020simclr} alongside other objectives in order to generate a more powerful visual representation for a given data distribution.


Given $N$ images in a minibatch, two random \textit{views} are generated for each image as a positive pair by random data augmentations, $\tau$ and $\tau'$. The remaining $2N-2$ augmented views of images within a minibatch are regarded as negative pairs. The augmentation comprises low-level image transformations. The contrastive feature of each view, $\mathbf{z}^C \in \mathbb{R}^{d^C}$, is produced by forwarding the pixel-wise features into the contrastive projection head, denoted as $c$. 

Then, we minimize the contrastive loss function for each data with the cosine similarity as follows:
\begin{equation}\label{simclr}
    \mathcal{L}_{simclr} = -log\frac{
        \exp(\frac{\mathbf{z}^C\cdot\mathbf{z}^C_{+}}{\lambda})}
        {
        \exp(\frac{\mathbf{z}^C \cdot \mathbf{z}^C_{+}}{\lambda}) + \sum_{\mathbf{z}^C_{-}}^{2N-2} \exp(\frac{\mathbf{z}^C \cdot \mathbf{z}^C_{-}}{\lambda})
        }
    .
\end{equation} $\lambda$ is a temperature hyperparameter, and $\mathbf{z}^C_{+}$ and $\mathbf{z}^C_{-}$ denotes features of positive and negative pairs, respectively. By minimizing the loss, features of positive pairs are pulled while pushing those of negative pairs away. In a self-supervised manner, it regularizes the features to be more semantically meaningful and discriminative for traversability estimation.


\section{Experiments}
In this section, we validate that our self-supervised traversability estimation method can effectively learn traversability in a wide range of environments. We first describe the dataset used for the evaluation, followed by the experimental setup as well as implementation details. Then, we present both quantitative and qualitative results of our traversability estimation method. Lastly, we present detailed ablation studies demonstrating that our self-supervised traversability estimation method is capable of learning traversability in a variety of environments and under appearance changes without human annotations.

\subsection{Datasets}\label{datasets}
\subsubsection{Driving Data Under Adverse Conditions}\label{taodac}
We collected driving data in a variety of environments using our platform~\cite{taodac} equipped with an RGB camera, VLP-32 LiDAR. It comprises about $20,000$ images gathered under a wide range of conditions, including varying places, seasons, weather, lighting, terrain types, obstacles, and lens conditions.

According to the characteristics of the environments, we divide our data into five categories: \textit{paved, unpaved, snowy, rainy}, and \textit{night}. The \textit{paved} contains images of urban and rural areas with paved roads. The \textit{unpaved} includes images acquired while driving on unpaved off-roads, where obstacles, dust, and smoke are captured on camera and the drivable regions are less clear. The \textit{snowy} and \textit{rainy} categories consist of images taken in the context of snow and rain, with snowed surfaces and puddles, as well as frost and raindrops on the lens. The \textit{night} is composed of images obtained in dark areas with headlights on. For evaluation, $300$ images per category are manually annotated by an expert. They are chosen from a subset of sequential images and excluded from training.

\subsubsection{RELLIS-3D}
We also present experimental results using the publicly available RELLIS-3D off-road dataset~\cite{Jiang_RELLIS3D}, which contains RGB camera images with pixel-level annotation. The self-supervised traversability labels are generated from the raw data using LiDAR-based SLAM~\cite{slam}. The data is divided into a training set containing $4,827$ images and a validation set with the remaining images. Although the annotation does not indicate which points are traversable, we define the \textit{grass, dirt, asphalt, concrete}, and \textit{mud} classes as traversable and the \textit{tree, pole, vehicle, object, person, fence, barrier, rubble}, and \textit{bush} classes as non-traversable.

\subsection{Experimental Setup}\label{setup}
First, we demonstrate that the model learned with our self-supervised traversability estimation methods is more effective than models trained in a fully supervised manner using datasets with human annotations. For the comparison, the PSPNet is trained in a supervised manner with the following datasets: \textit{KITTI} road detection, \textit{RELLIS-3D}, and our labeled dataset of outdoor driving scenes~(\textit{Outdoor}). For the outdoor dataset, about $1$K images of the paved and unpaved categories are randomly selected and manually labeled for supervised learning. The model is also trained with the aforementioned three datasets altogether~(ALL).

Then, we show that our method yields higher performances than other self-supervised traversability estimation algorithms. Our approach is compared with the method that uses normalizing flow~\cite{dinh_2017nvp} on top of the pre-trained backbone~(\textit{Real-NVP})~\cite{Wellhausen_2020} and the method based on reconstruction-based anomaly detection with autoencoder~(\textit{AE Based})~\cite{schmid_2022self}.

\subsection{Evaluation Metircs}\label{metrics}
The Area Under the Receiver Operating Characteristic~(AUROC) is used to quantitatively evaluate the methods. AUROC quantifies the likelihood of a positive sample having a higher normal score than a negative sample, and therefore evaluates the one-class classification algorithms regardless of the threshold. In addition, we evaluate our methods using the  standard evaluation metrics of the KITTI road detection system. Maximum F1-measure~(MaxF), average precision~(AP), precision rate~(PRE), recall rate~(REC), false positive rate~(FPR), and false negative rate~(FNR) are the metrics included. Note that the four latter measures are obtained at the threshold of the maximum F1 measure.

\subsection{Implementaion Details}\label{implementation}
We use PSPNet with ResNet50~\cite{he_2016ResNet} as a backbone embedding network for every method for fair comparisons. We use the flow model with eight transformation blocks composed of affine coupling layers. The contrastive head consists of adaptive average pooling and two MLP layers with a ReLU in the middle. Both the flow model and the contrastive head produce $128$ dimensional vectors that are $l_2$ normalized.

Our models are trained for $60$ epochs with a mini-batch size of $64$, using the Adam optimizer with a learning rate of $1e^{-3}$ and a polynomial learning rate decay. The sum of the means of the three losses~(Eq.~\ref{eq:svdd},\ref{CE}, and \ref{simclr}) is used as the objective of the optimization. The random data augmentation pipeline includes $256\times256$ pixel random cropping, flipping, random color jittering, random gray-scale conversion, gaussian blurring, rotation, and random perspective transformation. The number of learnable prototypes is set to $256$ and they are randomly initialized by the normal distribution. We execute three iterations of the Sinkhorn-Knopp algorithm and set the temperature parameters $\lambda$ as $0.1$. For data labeling, we set horizons $\alpha$ for self-supervised labels in \Sref{sec:label} to $100$, indicating that we utilized trajectories $10$ seconds ahead of the image acquisition.

\subsection{Experimental Results}\label{results}

\begin{table*}[t]
\centering
\renewcommand{\arraystretch}{1.5}
\caption{Quantitative results of traversability estimation methods for our dataset collected under various conditions.}
\label{tab:result}
\resizebox{0.9\textwidth}{!}{%
\centering
\begin{tabular}{cccccccccccccccc}
\toprule
 &
  \multicolumn{3}{c}{Paved} &
  \multicolumn{3}{c}{Unpaved} &
  \multicolumn{3}{c}{Snowy} &
  \multicolumn{3}{c}{Rainy} &
  \multicolumn{3}{c}{Night} \\ \cmidrule(lr){2-4} \cmidrule(lr){5-7} \cmidrule(lr){8-10} \cmidrule(lr){11-13} \cmidrule(lr){14-16}  
\multicolumn{1}{c}{\multirow{-2}{*}{}} &
  AUROC &
  MaxF &
  AvgPrec &
  AUROC &
  MaxF &
  AvgPrec &
  AUROC &
  MaxF &
  AvgPrec &
  AUROC &
  MaxF &
  AvgPrec &
  AUROC &
  MaxF &
  AvgPrec \\ \midrule\midrule    
\multicolumn{1}{l|}{PSPNet (KITTI~\cite{geiger2012_kitti})}& 0.8732 & 0.7820 & 0.8570 & 0.9100 & 0.8039 & 0.8486 & 0.6669 & 0.7467 & 0.8245 & 0.9095 & 0.8034 & 0.8605 & 0.9323 & 0.8778 & 0.9256 \\
\multicolumn{1}{l|}{PSPNet (RELLIS-3D~\cite{Jiang_RELLIS3D})} & 0.9143 & 0.7728 & 0.7562 & 0.9401 & 0.8398 & 0.7690 & 0.6187 & 0.7670 & 0.7223 & 0.9143 & 0.7728 & 0.7562 & 0.9669 & 0.9168 & 0.9059 \\
\multicolumn{1}{l|}{PSPNet (Outdoor)}& 0.9861 & \textbf{0.9414} & \textbf{0.9386} & 0.9729 & 0.8829 & 0.8828 & 0.7592 & 0.8593 & 0.8945 & 0.9757 & 0.8899 & 0.9094 & 0.9673 & 0.9139 & 0.9372 \\ 
\multicolumn{1}{l|}{PSPNet (ALL)}& \textbf{0.9870} & 0.9396 & 0.9383 & 0.9720 & 0.9051 & 0.8809 & 0.9459 & 0.8337 & 0.8454 & 0.9672 & 0.8839 & 0.9019 & 0.9565 & 0.9105 & 0.9338 \\ \cdashline{1-16}
\multicolumn{1}{l|}{Real-NVP~\cite{Wellhausen_2020}}& 0.5878 & 0.6101 & 0.4863 & 0.6352 & 0.5075 & 0.4313 & 0.5459 & 0.5245 & 0.3719 & 0.6712 & 0.5175 & 0.4878 & 0.5145 & 0.6286 & 0.5344 \\
\multicolumn{1}{l|}{AE Based~\cite{schmid_2022self}}& 0.7923 & 0.7102 & 0.6905 & 0.8443 & 0.6795 & 0.6712 & 0.8120 & 0.6683 & 0.6544 & 0.8321 & 0.6600 & 0.6677 & 0.8609 & 0.7851 & 0.8235 \\
\rowcolor[HTML]{C0C0C0}\multicolumn{1}{l|}{\textbf{Ours}}& 0.9660 & 0.8888 & 0.9306 & \textbf{0.9813} & \textbf{0.9143} & \textbf{0.9168} & \textbf{0.9770} & \textbf{0.8930} & \textbf{0.9052} & \textbf{0.9815} & \textbf{0.9044} & \textbf{0.9160} & \textbf{0.9750} & \textbf{0.9169} & \textbf{0.9387} \\  \bottomrule
\end{tabular}
}
\end{table*}

\begin{figure*}[t]
\centering
\includegraphics[width=0.9\linewidth]{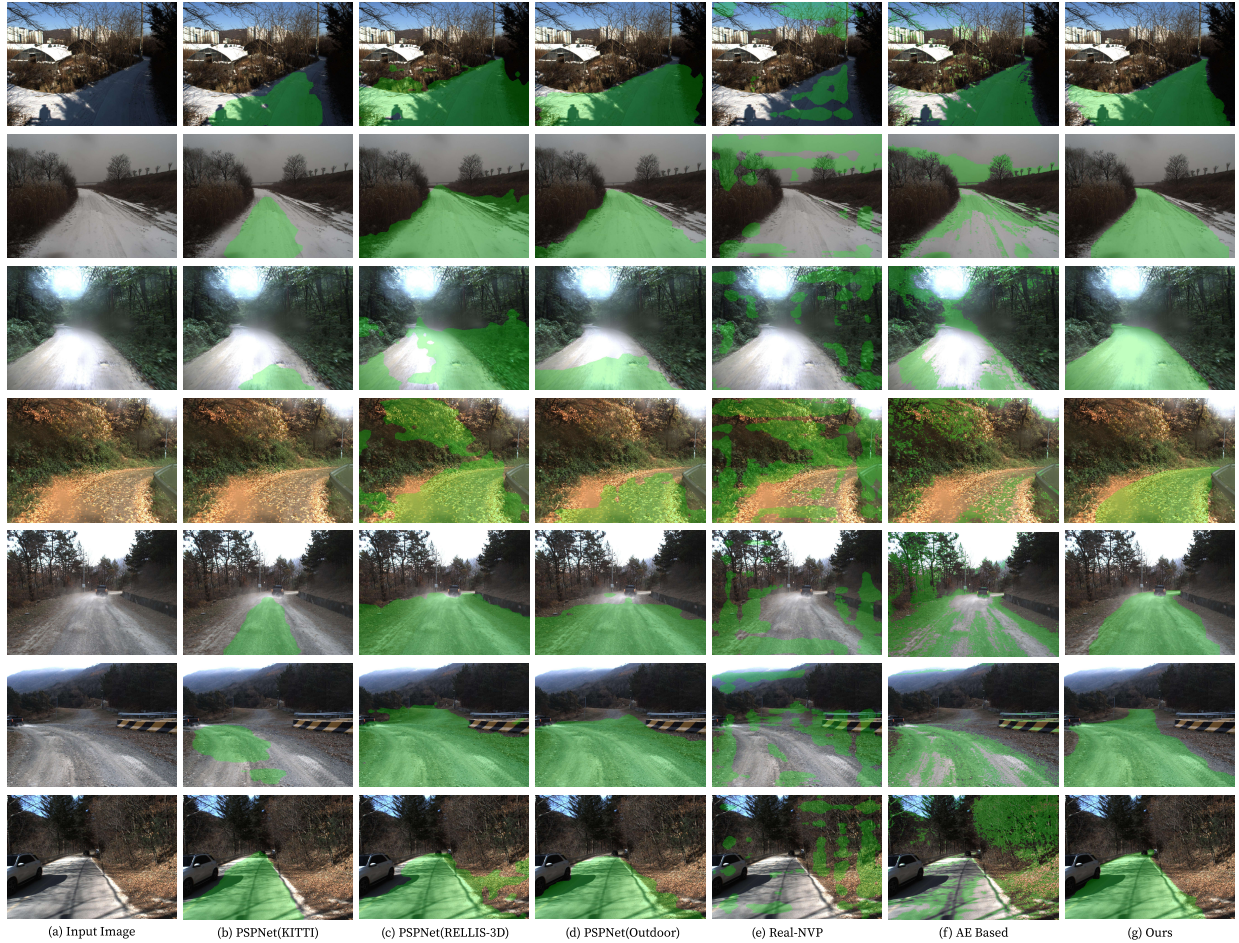}
\caption{Qualitative traversability estimation results for our dataset. The thresholds of the maximum F1 scores are used for the visualizations. The pixels where the estimated traversability exceeds the thresholds are colored green. More results are available in the multimedia material.
}
\label{fig:MAIN_RESULTS}
\end{figure*}

The \Tref{tab:result} and \Fref{fig:MAIN_RESULTS} provide the quantitative and qualitative results for our dataset. In most of the categories, our method outperforms models trained with manual labels in a supervised way, implying that our methods can yield comparable or even greater results than a model trained with laborious manual annotations. In addition, combining multiple datasets for supervised learning does not seem to improve the traversability estimation performance on target distributions. These results demonstrate that distribution shifts have a significant impact on the performance of traversability estimation, suggesting the necessity of self-supervised traversability estimation methods for autonomous vehicles to operate effectively in widespread environments. Due to the fact that our method exploits self-supervised labels of the target distributions, the model can estimate traversability reliably in the presence of distributional shifts.

Our method shows a significant margin compared to other self-supervised methods based on one-class classification. Note that the methods based on autoencoders produce a large number of false positives, meaning that the simple reconstruction-based anomaly detections fail to distinguish non-traversable regions. Ours produces fewer false negative occurrences than others, indicating that it obtains discriminative features for identifying traversable regions in complex off-road environments.

\begin{figure}[t]
\centering
\includegraphics[width=0.9\linewidth]{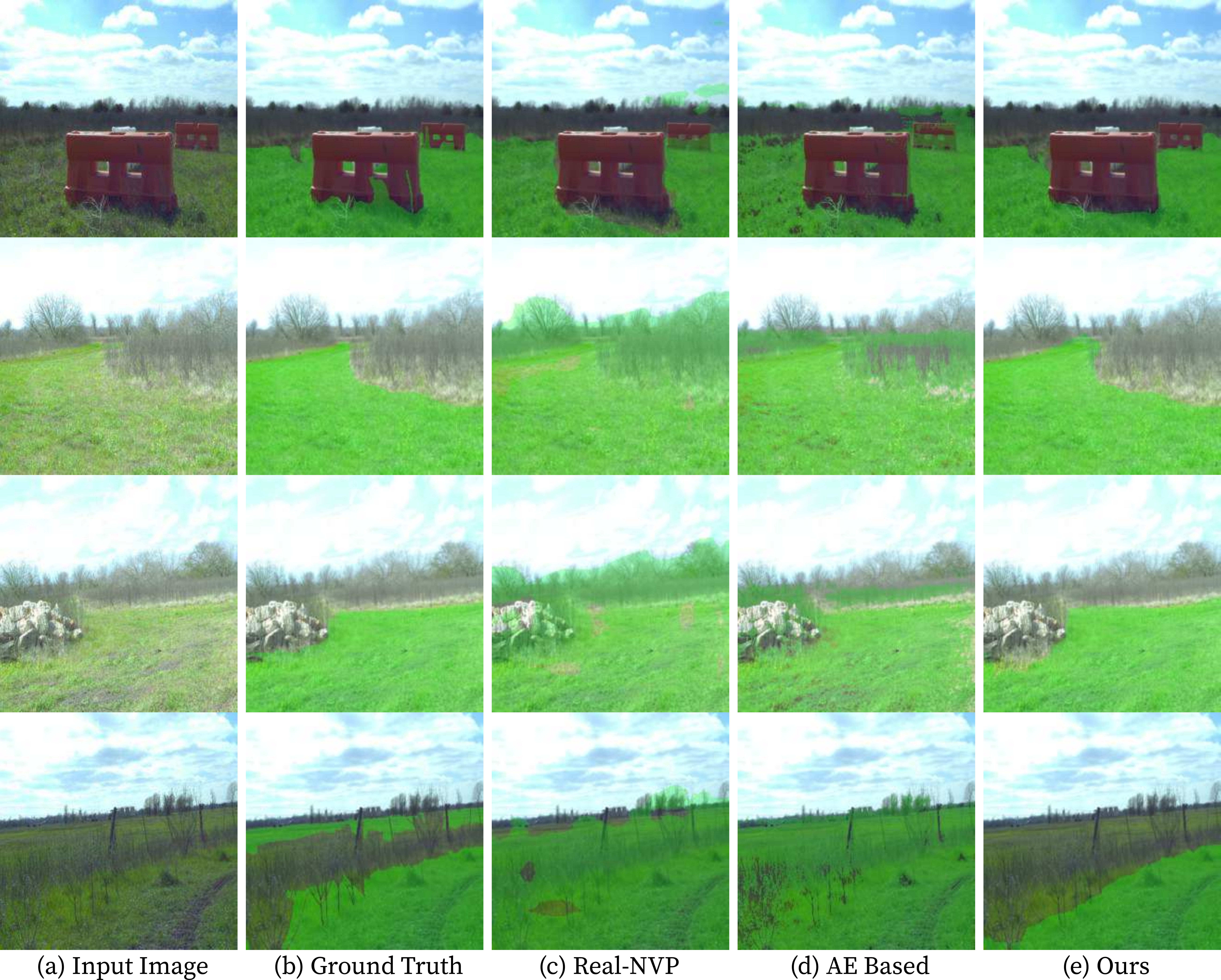}
\caption{Qualitative results for RELLIS-3D. Note that the ground truths are sometimes inaccurate and labeled for terrain classification rather than traversability. For example, some pixels on obstacles are labeled as traversable due to thin overhanging vegetation, and distant grass across fences is also labeled as traversable. Instead of merely assessing terrain type, our method identifies traversable regions that are contextually significant.
}
\label{fig:RELLIS_RESULTS}
\end{figure}

The \Tref{tab:results_rellis3d} and \Fref{fig:RELLIS_RESULTS} illustrate results for RELLIS-3D. Our method shows better performance than others and even yields comparable performance with models overfit with manual annotations of the RELLIS-3D.

\subsection{Ablation Studies}\label{ablations}
\begin{table}[t]
\centering
\renewcommand{\arraystretch}{1.5}
\caption{Quantitative results for RELLIS-3D.}
\label{tab:results_rellis3d}
\resizebox{0.9\linewidth}{!}{%
\begin{tabular}{rccccccc}
\toprule
\multicolumn{1}{l}{}  & AUROC & MaxF & AvgPrec & PRE & REC & FPR & FNR \\ \midrule \midrule
\multicolumn{1}{l|}{PSPNet (KITTI~\cite{geiger2012_kitti})}& 0.7565 & 0.7345 & 0.7431 & 0.6241 & 0.8922 & 0.6248 & 0.1078\\
\multicolumn{1}{l|}{PSPNet (RELLIS-3D~\cite{Jiang_RELLIS3D})} & \textbf{0.9814} & \textbf{0.9400} & \textbf{0.9457} & 0.9378 & \textbf{0.9423} & 0.0702 & \textbf{0.0578} \\
\multicolumn{1}{l|}{PSPNet (Outdoor)}& 0.8160 & 0.7860 & 0.7895 & 0.7104 & 0.8798 & 0.4030 & 0.1202 \\ 
\multicolumn{1}{l|}{PSPNet (ALL)} & 0.9814 & 0.9396 & 0.9454 & \textbf{0.9388} & 0.9404 & \textbf{0.0690} & 0.0607 \\\cdashline{1-8}
\multicolumn{1}{l|}{Real-NVP~\cite{Wellhausen_2020}}& 0.5625 & 0.7001 & 0.5710 & 0.5464 & 0.9742 & 0.9280 & 0.0258 \\
\multicolumn{1}{l|}{AE Based~\cite{schmid_2022self}}& 0.7348 & 0.7437 & 0.7079 & 0.6250 & 0.9181 & 0.6323 & 0.0819 \\
\rowcolor[HTML]{C0C0C0}\multicolumn{1}{l|}{\textbf{Ours}} & 0.9036 & 0.8622 & 0.9164 & 0.8738 & 0.8508 & 0.1466 & 0.1492 \\ 
\bottomrule
\end{tabular}%
}
\end{table}

\begin{figure}[t]
\centering
\includegraphics[width=0.95\linewidth]{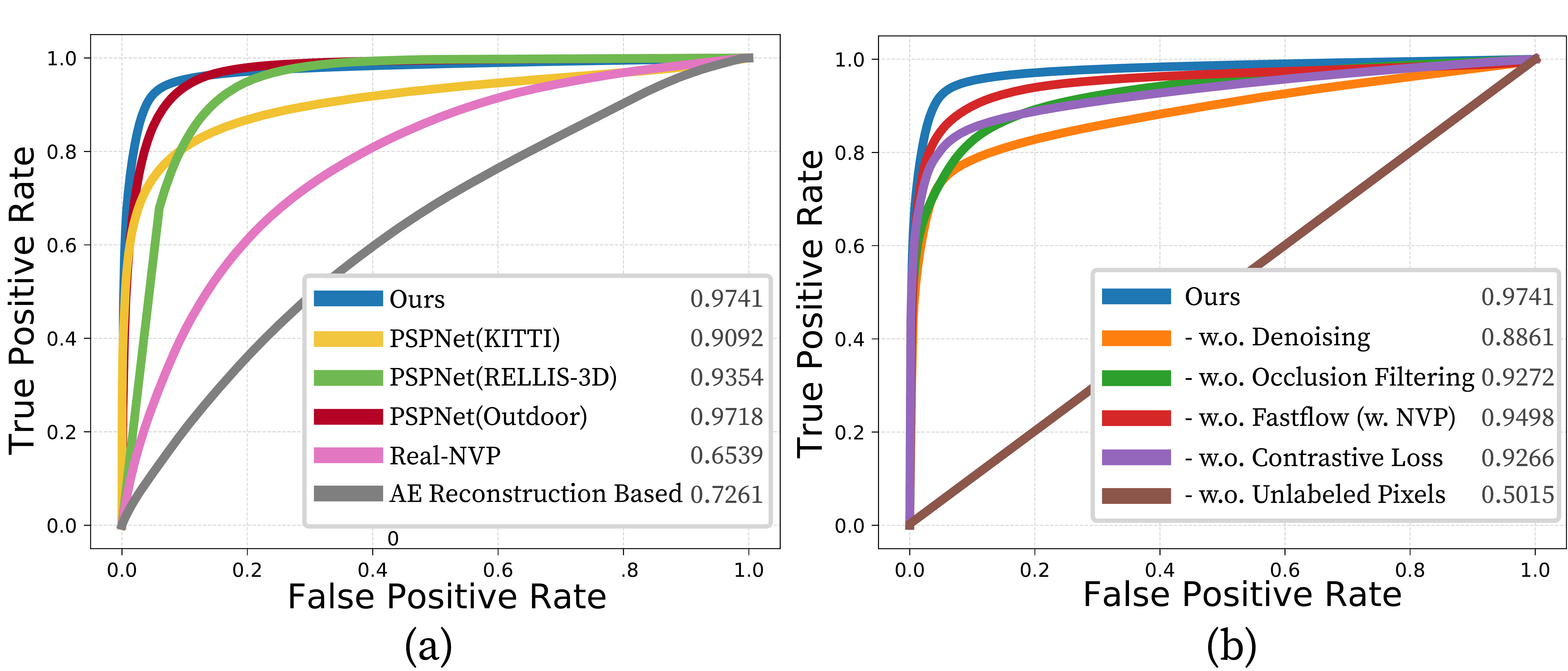}
\caption{ROC curves for (a) traversability estimation methods and (b) the ablation studies. The values indicate the AUROC of the results. Notably, our method yields low false positive rates compared to other methods, which is essential for safe navigation in unstructured environments.}
\label{fig:ROC}
\end{figure}

We present comprehensive ablation studies to examine the validity of each component of our methodology. We quantitatively verify the efficacy of the self-supervised traversability labeling, PU learning algorithm with the normalizing flow, self-supervised clustering, and self-supervised contrastive learning. The ablations are trained using data from all categories of our dataset. The results are shown in \Tref{tab:ablation} and \Fref{fig:ROC} shows the ROC curves of the results.

\begin{table}[t]
\centering
\renewcommand{\arraystretch}{1.5}
\caption{Quantitative results of the ablation studies.}
\label{tab:ablation}
\resizebox{0.9\linewidth}{!}{%
\begin{tabular}{rccccccc}
\toprule
\multicolumn{1}{l}{}                                        & AUROC & MaxF & AvgPrec & PRE & REC & FPR & FNR \\ \midrule \midrule
\multicolumn{1}{l|}{PSPNet (KITTI~\cite{geiger2012_kitti})}                                  & 0.9092 & 0.8044 & 0.8504 & 0.8451 & 0.7674 & 0.0643 & 0.2326 \\
\multicolumn{1}{l|}{PSPNet (RELLIS-3D~\cite{Jiang_RELLIS3D})}                              & 0.9354 & 0.8126 & 0.7644 & 0.7521 & 0.8836 & 0.1331 & 0.1164 \\
\multicolumn{1}{l|}{PSPNet (Outdoor)}                                & 0.9718 & 0.8765 & 0.8927 & 0.8532 & 0.9011 & 0.0708 & 0.0989 \\
\multicolumn{1}{l|}{PSPNet (ALL)}                                & 0.9732 & 0.9013 & 0.8909 & 0.8806 & 0.9230 & 0.0572 & 0.0770 \\
\cdashline{1-8}
\multicolumn{1}{l|}{Real-NVP~\cite{Wellhausen_2020}}                               & 0.6539 & 0.5326 & 0.4480 & 0.3838 & 0.8698 & 0.6856 & 0.1302 \\
\multicolumn{1}{l|}{AE Based~\cite{schmid_2022self}}                               & 0.7261 & 0.5847 & 0.5342 & 0.4779 & 0.7530 & 0.4039 & 0.2470 \\ 
\rowcolor[HTML]{C0C0C0}\multicolumn{1}{l|}{\textbf{Ours}} & \textbf{0.9741} & \textbf{0.9082} & \textbf{0.9143} & \textbf{0.9014} & \textbf{0.9151} & \textbf{0.0456} & \textbf{0.0849} \\ \cdashline{1-8}
\rowcolor[HTML]{E6E6E6}\multicolumn{1}{l|}{1) Self-Supervised Labels} &&&&&&&\\
\rowcolor[HTML]{D1D1D1}\multicolumn{1}{r|}{w.o. Occlusion Filtering}               & 0.9272 & 0.8144 & 0.8682 & 0.8201 & 0.8087 & 0.0871 & 0.1913 \\
\rowcolor[HTML]{D1D1D1}\multicolumn{1}{r|}{w.o. Denoising}                         & 0.8861 & 0.8019 & 0.8414 & 0.8679 & 0.7452 & 0.0557 & 0.2548 \\ 
\rowcolor[HTML]{D1D1D1}\multicolumn{1}{r|}{w. horizon $\alpha=30$}               & 0.9497 & 0.8839 & 0.9125 & 0.8846 & 0.8833 & 0.0530 & 0.1167 \\
\rowcolor[HTML]{D1D1D1}\multicolumn{1}{r|}{w. horizon $\alpha=50$}               & 0.9632 & 0.8861 & 0.90951 & 0.8841 & 0.8880 & 0.0535 & 0.1120 \\
\cdashline{1-8}
\rowcolor[HTML]{E6E6E6}\multicolumn{1}{l|}{2) PU Learning} &&&&&&&\\
\rowcolor[HTML]{D1D1D1}\multicolumn{1}{r|}{w.o. Fastflow(w. Real-NVP)}                  & 0.9498 & 0.8690 & 0.8994 & 0.8852 & 0.8534 & 0.0544 & 0.1466 \\
\rowcolor[HTML]{D1D1D1}\multicolumn{1}{r|}{w.o. Fastflow(w. MLP)}                  & 0.5948 & 0.5236 & 0.5049 & 0.3555 & 0.9934 & 0.8843 & 0.0066 \\ 
\rowcolor[HTML]{D1D1D1}\multicolumn{1}{r|}{w.o. Unlabeled Pixels}                  & 0.5015 & 0.4954 & 0.3879 & 0.3293 & 1.0000 & 1.0000 & 0.0000 \\ 
\cdashline{1-8}
\rowcolor[HTML]{E6E6E6}\multicolumn{1}{l|}{3) Self-Supervised Learning} &&&&&&&\\
\rowcolor[HTML]{D1D1D1}\multicolumn{1}{r|}{w.o. Contrastive Loss}                  & 0.9266 & 0.8449 & 0.8829 & 0.8831 & 0.8098 & 0.0526 & 0.1902 \\
\rowcolor[HTML]{D1D1D1}\multicolumn{1}{r|}{w. Reconstruction Loss}                 & 0.9348 & 0.8191 & 0.8684 & 0.8083 & 0.8301 & 0.0966 & 0.1699 \\ 
\bottomrule
\end{tabular}%
}
\end{table}

\subsubsection{Self-Supervised Labels}
First, to validate the efficacy of our self-supervised labeling algorithm, we compare the impacts of labels obtained without occlusion filtering, without LiDAR denoising, and with varying horizons. The labels generated without occlusion filtering result in a high FPR, indicating that non-navigable regions, such as obstacles, are incorrectly estimated as traversable due to unreliable labels. The labels created without LiDAR denoising lead to a higher FNR, implying that the model is trained with fewer supervisions because self-labels for traversable regions are misclassified as occluded due to LiDAR noise. Similarly, lowering the horizons of trajectories increased the FNR because it reduces the number of positive pixels of the labels.

\subsubsection{PU Learning}
Second, we replace the flow model with $1D$ flow~(Real-NVP) and simple MLP layers to evaluate the efficacy of flow models. We observe that $2D$ normalizing flow produces better results compared to $1D$ flow. It confirms that using $2D$ normalizing flow is more effective for localizing traversable regions with pixel-wise features while avoiding a trivial solution. The model trained by replacing the flow model with MLPs results in low AUROC and high FPR, implying that the models  produce a trivial solution without the normalizing flow. Also, the model is trained without clustering loss in order to highlight the effectiveness of using unlabeled data. Without the loss for unlabeled pixels, \Eref{CE}, the performance severely diminishes as AUROC approaches $0.5$. It indicates that the use of unlabeled data is essential for avoiding trivial solutions and learning more discriminative features about environments.

\subsubsection{Self-Supervised Learning for Visual Representation}
Then, we verify the efficacy of our self-supervised contrastive learning. The models are trained without contrastive learning and with the reconstruction pretext task. The model trained without contrastive learning exhibits a low REC, denoting that the image features are less relevant for learning traversability, as the self-labels cannot provide guidance regarding the distinction between traversable and non-traversable regions. The performance of the model learned concurrently with the reconstruction pretext task is improved, but still inferior to the model with contrastive loss. The reason for this is that the naive reconstruction objective tends to focus on texture and color rather than semantic meanings.


\section{Conclusions}


This paper introduces a self-supervised traversability estimation method that can learn traversability in varied environments without manual annotations. Using the self-supervised labels only, the network is trained to predict traversable regions using a one-class classification method. Self-supervised learning of visual representation is incorporated into the learning in order to improve the network. Extensive experiments demonstrate that the proposed method is capable of learning traversability more effectively than others. We believe that our method can be leveraged for the wider deployment of autonomous vehicles since it is capable of easily adapting to a variety of contexts by precisely embedding the data distribution of target environments. 

Future work includes using labeled data with domain adaptation and semi-supervised learning. Also, we are investigating incremental learning and online learning for a more general model that can be used in various environments.

\addtolength{\textheight}{0cm}   

\bibliographystyle{IEEEtran}
\bibliography{mybib.bib}

\end{document}